\documentclass[runningheads]{llncs}
\usepackage{graphicx}
\usepackage{booktabs}
\usepackage{amssymb}
\usepackage{amsmath}
\usepackage{pifont}
\usepackage{colortbl}
\usepackage{latexsym}
\usepackage{multirow}
\usepackage{xspace}

\usepackage[hidelinks]{hyperref}
\usepackage[capitalize]{cleveref}
\crefname{section}{Sec.}{Secs.}
\Crefname{section}{Section}{Sections}
\Crefname{table}{Table}{Tables}
\crefname{table}{Tab.}{Tabs.}
\renewcommand\UrlFont{\color{blue}\rmfamily}
\hypersetup{
    colorlinks=true,
    linkcolor=blue,
    urlcolor=blue,
    citecolor=blue
}
\newcommand{\mpm}{\textbf{MPM}}
\newcommand{\stage}[1]{stage~\uppercase\expandafter{\romannumeral#1}}
\newcommand{\Stage}[1]{Stage~\uppercase\expandafter{\romannumeral#1}}

\makeatletter
\DeclareRobustCommand\onedot{\futurelet\@let@token\@onedot}
\def\@onedot{\ifx\@let@token.\else.\null\fi\xspace}

\def\eg{\emph{e.g}\onedot}

\def\etc{\emph{etc}\onedot}

\def\etal{\emph{et al}\onedot}
\usepackage{marvosym}
\renewcommand\UrlFont{\color{blue}\rmfamily}
\begin{document}
\title{MPM: A Unified 2D-3D Human Pose Representation via Masked Pose Modeling}
\titlerunning{MPM}
\renewcommand\UrlFont{\color{blue}\rmfamily}

\author{ Zhenyu Zhang\inst{1,2,3}\ \and
Wenhao Chai\inst{4} \and
Zhongyu Jiang\inst{4}\and 
Tian Ye\inst{5}\and
Mingli Song\inst{1,2,3}\and
Jenq-Neng Hwang\inst{4}\and
Gaoang Wang\inst{6}\textsuperscript{(\Letter)}
}
\institute{Anonymous Institute}
\authorrunning{Zhang et al.}
\institute{State Key Laboratory of Blockchain and Data Security, Zhejiang University, Hangzhou, China\\\and
 Key Laboratory of Visual Perception (Zhejiang University), Ministry of Education and Microsoft, Hangzhou, China\and
 Hangzhou High-Tech Zone (Binjiang) Institute of Blockchain and Data Security, Hangzhou, China  \\
 \email{\{zhenyuzhang, brooksong\}@zju.edu.cn}\and
University of Washington, Seattle, USA\\ \email{ \{wchai, zyjiang, hwang\}@uw.edu}\and
The Hong Kong University of Science and Technology (Guangzhou), Guangzhou, China\\ \email{owentianye@hkust-gz.edu.cn} \and
ZJU-UIUC Institute, Zhejiang University, Haining, China\\ \email{gaoangwang@intl.zju.edu.cn}}

% \textbf{Zhenyu Zhang}$^{1}$ \footnotemark[1] \quad \textbf{Wenhao Chai}$^{2}$ \thanks{~Equal contribution. $^\dagger$ Corresponding author.} \quad \textbf{Zhongyu Jiang}$^2$ \quad \textbf{Tian Ye}$^3$ \\
% \textbf{Mingli Song}$^1$ \quad \textbf{Jenq-Neng Hwang}$^2$ \quad \textbf{Gaoang Wang}$^{4\dagger}$\\
% Zhejiang University$^1$ \quad University of Washington$^2$ \\
% Jimei University$^3$ \quad ZJU-UIUC Institute, Zhejiang University$^4$\\
% \tt\small \{zhenyuzhang, brooksong\}@zju.edu.cn, \{wchai, zyjiang, hwang\}@uw.edu, \\
% \tt\small 201921114031@jmu.edu.cn, gaoangwang@intl.zju.edu.cn

\maketitle              % typeset the header of the contribution
\makeatother

\begin{abstract}
Estimating 3D human poses only from a 2D human pose sequence is thoroughly explored in recent years. Yet, prior to this, no such work has attempted to unify 2D and 3D pose representations in the shared feature space. In this paper, we propose \mpm, a unified 2D-3D human pose representation framework via masked pose modeling. We treat 2D and 3D poses as two different modalities like vision and language and build a single-stream transformer-based architecture. We apply two pretext tasks, which are masked 2D pose modeling, and masked 3D pose modeling to pre-train our network and use full-supervision to perform further fine-tuning. A high masking ratio of $71.8~\%$ in total with a spatio-temporal mask sampling strategy leads to better relation modeling both in spatial and temporal domains. \mpm~can handle multiple tasks including 3D human pose estimation, 3D pose estimation from occluded 2D pose, and 3D pose completion in a \textbf{single} framework. We conduct extensive experiments and ablation studies on several widely used human pose datasets and achieve state-of-the-art performance on MPI-INF-3DHP. 
\end{abstract}

\vspace{-10pt}
\begin{figure}[t]
    \centering
    \includegraphics[width=0.7\linewidth]{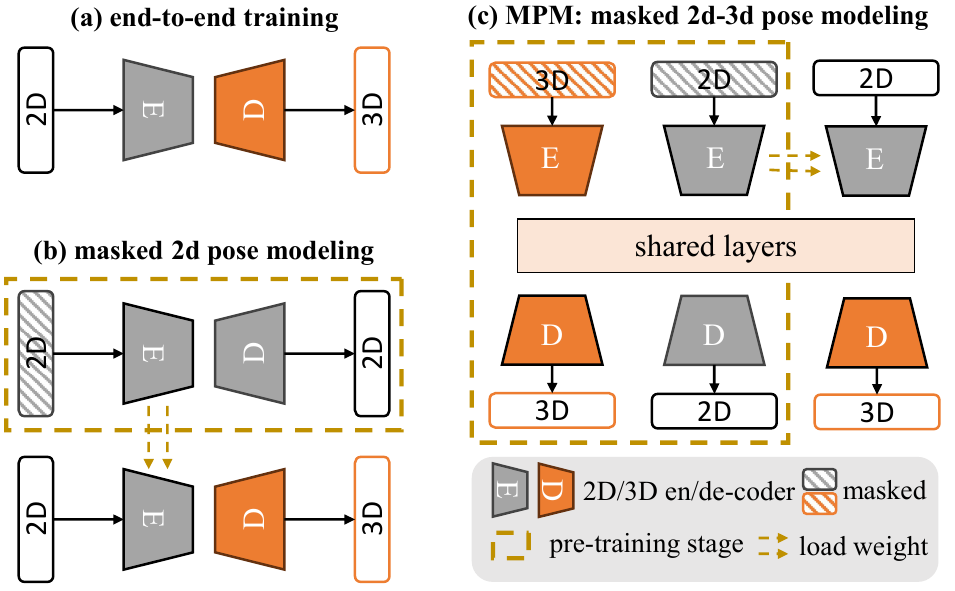}
    \caption{Comparison of the existing 3d human pose estimation lifting method and our \mpm. (a) End-to-end training without pre-training using any backbone; (b) Pre-training with mask 2D pose modeling and then fine-tuning~\cite{shan2022p}; (c) our \mpm: pre-trained with both mask 2D and 3D pose modeling, in which most of the parameters in shared layers to learn a unified representation.}
    \label{fig:intro}
\vspace{-8pt}
\end{figure}

\section{Introduction}
3D human pose estimation (HPE) tasks aim to estimate 3D human joint positions from a single image or a video clip. 3D HPE has a wide range of applications in various computer vision tasks, such as action recognition~\cite{li2021symbiotic,yang2021feedback}, multiple object tracking~\cite{hao2023divotrack}, human parsing~\cite{chen20173d,chen2018cascaded}, the so-called lifting paradigm. Lifting methods fully utilize the image/video-2D pose datasets (\eg,COCO~\cite{lin2014microsoft}) and 2D-3D pose datasets (\eg,Human3.6M~\cite{ionescu2013human3})  to form the two-stage methods. Yet, existing 2D-3D pose datasets either lack enough diversity in laboratorial environments~\cite{ionescu2013human3} or lack adequate quantity and accuracy in the wild~\cite{mehta2017monocular}, resulting in a large domain gap among different datasets and also in real-world scenarios. There are few prior arts aiming to improve the generalization~\cite{gong2021poseaug} or adaptation~\cite{chai2023global} capability among various domains. Nonetheless, developing a human pose pre-training paradigm by utilizing all the 2D-3D pose datasets is far from fully explored.

As for the lifting network architectures, early works are based on fully-connected layers networks~\cite{martinez2017simple}, 3D convolution networks~\cite{pavllo20193d}, and graph neural networks~\cite{yan2018spatial}. Recently, transformer-based~\cite{vaswani2017attention} methods also have achieved state-of-the-art performance in 3D human pose estimation tasks~\cite{zheng20213d,zhao2023poseformerv2,li2022mhformer} following lifting paradigm. Empirically, the amount of data used to train the model with transformer architecture is often much more than those with other backbones. On the other hand, the pre-training paradigm in transformers is widely discussed, which might also benefit human pose estimation tasks. Inspired by some previously proposed self-supervised pre-training methods like masked modeling~\cite{shan2022p} and also the transformer architecture design in multi-modality area~\cite{luo2020univl}, we propose \mpm, a unified 2D-3D human pose representation network pre-trained by masked pose modeling paradigm.

We treat 2D and 3D pose as two different modalities like vision and language and build a single-stream transformer-based architecture. To be specific, as shown in Figure~\ref{fig:intro}, we first use two separate encoders to embed 2D and 3D poses into a shared embedding space. After that, the 2D or 3D pose embedding is put into shared transformer layers, which contain most of the parameters of our network. Note that only one of the 2D and 3D poses are put in at once. Finally, two separate decoders are used to decode 2D and 3D poses from embedding.

We use two training stages including masked modeling pre-training (\stage{1}) and full-supervised fine-tuning (\stage{2}) to train our network. In \stage{1}, we apply two pretext tasks including (1) masked 2D pose modeling, (2) masked 3D pose modeling. Note that we conduct masked operations both in spatial and temporal directions. After pre-training on those masked modeling pretext tasks, the model has learned the prior knowledge of spatial and temporal relations of both 2D and 3D human pose information between them. In \stage{2}, we use 2D-3D pose pairs to perform fine-tuning on 3D HPE task since the network is trained with masked pose input, which causes the gap when unmasked poses are set as input. We also use partial 2D-3D and partial 3D-3D pose pairs to further explore pose completion tasks.

Our contributions are summarized as follows:
% \vspace{-5pt}
\begin{itemize}

\item We are the first to treat 2D and 3D pose as two different modalities in a shared embedding space and build a single-stream transformer-based architecture for 3D human pose estimation and pose completion tasks.

\item We apply two masked modeling based pretext tasks for human pose pre-training to learn spatial and temporal relations.

\item We conduct extensive experiments and ablation studies on multiple widely used human pose datasets and achieve state-of-the-art performance on MPI-INF-3DHP benchmark.

\end{itemize}
\section{Related Works}
\subsection{3D Human Pose Estimation in Video}

Estimating 3D human pose only from 2D human pose sequences is the common paradigm for 3D human pose estimation in video and has been thoroughly explored in recent years. Pavllo~\etal ~\cite{pavllo20193d} leverage dilated temporal convolutions with semi-supervised way to improve 3D pose estimation in videos. Martinez~\etal~\cite{martinez2017simple} propose an MLP-based network for lifting 2D to 3D poses. Transformer-based networks achieve state-of-the-art performance recently. Zheng~\etal~\cite{zheng20213d} is the first to introduce a transformer into a 3D human pose estimation task. Li~\etal~\cite{li2022mhformer} propose a multi-hypothesis transformer that learns spatio-temporal representations of multiple plausible pose hypotheses. Those prior arts show the effectiveness of transformer architectures for 3D human pose estimation tasks in video. We take successful experiences to design our network.

\subsection{Transformer-based Multimodal Architecture}
As we treat 2D and 3D poses as two different modalities, we also learn from the transformer-based architecture from multimodal~\cite{xu2023multimodal} field. According to the network structures, a multimodal transformer can be divided into single-stream (\eg, Uniter~\cite{chen2020uniter}), multi-stream (\eg, ViLBERT~\cite{lu2019vilbert}) \etc. To learn a unified 2D/3D human pose representation, we follow the two-stream architecture design, in which a large amount of the parameters are shared between 2D and 3D poses. The encoders and decoders are relatively light compared to the shared transformer layers. In this case, we hope the prior knowledge can be maximized between 2D and 3D human poses in spatial and temporal domains.

\subsection{Network Pre-training via Masked Modeling}
Masked modeling as the pre-training task is first used in the natural language processing (NLP) field. BERT~\cite{devlin2018bert} masks out a portion of the input language tokens and trains models to predict the missing content. In the computer vision (CV) field, MAE~\cite{he2022masked} also shows the effectiveness of mask modeling by masking random patches of the input image and reconstructing the missing pixels. P-STMO~\cite{shan2022p} is the first to introduce masked pose modeling in the 3D human pose estimation task. It randomly masks some of the 2D poses, and reconstructs the complete sequences with a model consisting of spatial encoder, temporal encoder, and decoder. Yet, P-STMO has not explored masked 3D pose modeling as well as the unified representation of 2D and 3D human pose. Besides, pre-training with multiple 2D-3D pose datasets are not explored either. In this paper, we show that 3D poses can also be exploited in pre-training tasks in the same way as 2D and even further in a unified manner. 
\begin{figure}[!t]
    \centering
    \includegraphics[width=0.8\linewidth]{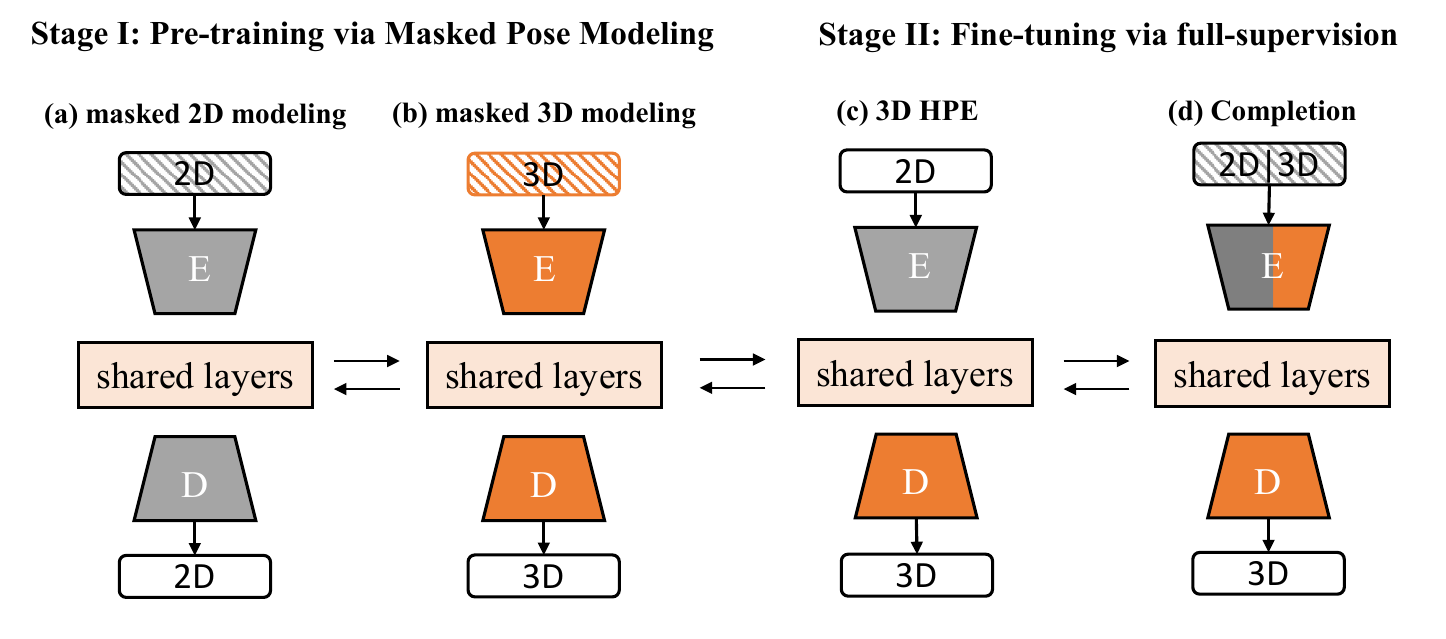}
    \caption{We conduct two training stages to train the proposed network. In \stage{1}, we apply two pretext tasks including (a) masked 2D modeling, (b) masked 3D modeling. After pre-training on those masked modeling pretext tasks, the model has learned the prior knowledge of spatial and temporal relations of both 2D and 3D human poses. In \stage{2}, we use unmasked 2D-3D pose pairs to further fine-tune the network on 3D HPE or use partial 2D-3D/partial 3D-3D to perform fine-tune on pose completion.}
    \label{fig:method}
\end{figure}

\section{Methodology}
\subsection{Architecture}
\subsubsection{2D/3D encoder.}
We use a simple MLP as the architecture for both the 2D and 3D pose encoders, and use 1D convolution with kernel size 1 as a fully connected layer. 2D and 3D poses are embedded into the same-dimension features. Note that we encode the input pose sequence \textit{frame-by-frame}, which has no temporal relation modeling. The encoders are relatively light as shown in Table~\ref{tab:param} compared to shared transformer layers since we aim to force 2D and 3D poses to share the unified representation. This process can be formulated as:
\begin{equation}
    \mathrm{f}^0_{2D} = \mathcal{E}_{2D}(\mathrm{P}^{in}_{2D}),\quad \mathrm{f}^0_{3D} = \mathcal{E}_{3D}(\mathrm{P}^{in}_{3D}),
\end{equation}
where $\mathrm{P}^{in}_{2D} \in \mathbb{R}^{l \times J \times 2}$ and $\mathrm{P}^{in}_{3D}$ are the input 2D and 3D pose sequences, $\mathcal{E}_{2D}(\cdot)$ and $\mathcal{E}_{3D}(\cdot)$ are 2D and 3D encoders, and $\mathrm{f}^0_{2D}$ and $\mathrm{f}^0_{3D}$ are 2D and 3D pose embeddings before being put in shared layers which share the common space.

\subsubsection{Shared layers.}
In shared layers, we use both MLP and transformer-based architectures to process the information on pose sequences. Shared layers contain the most parameters of our network.

\begin{figure}[t]
    \centering
    \includegraphics[width=0.8\linewidth]{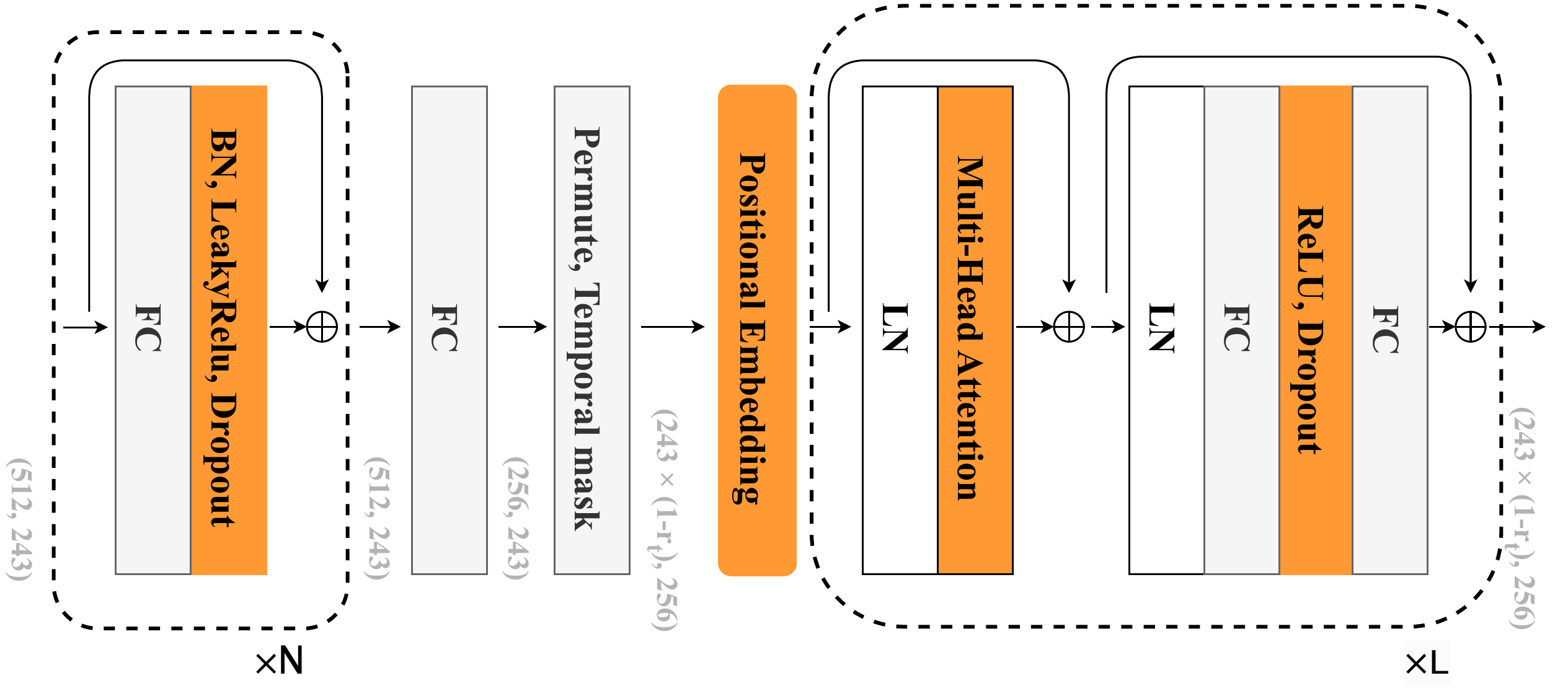}
    \caption{Detailed architecture of the shared transformer layers. We take $243$ frames as the tensor size in the pipeline.}
    \label{fig:transformer}
\end{figure}

We model spatial relations through a shared MLP, which is much heavier than the encoders, and model temporal relations through shared temporal transformer layers as shown in Figure~\ref{fig:transformer}. The pose embedding input is first calculated by the MLP-based spatial encoder. This process can be formulated as: 
\begin{equation}
    \mathrm{f}^n_{2D} = \mathcal{S}(\mathrm{f}^0_{2D}),\quad \mathrm{f}^n_{3D} = \mathcal{S}(\mathrm{f}^0_{3D}),
\end{equation}
where $\mathcal{S}(\cdot)$ is the shared spatial layers. 

The shared temporal layers are based on a vanilla transformer architecture as~\cite{shan2022p}. We treat frames as tokens to model the temporal relation of poses in the same sequence. 
The process can be formulated as: 
\begin{equation}
    \mathrm{f}^l_{2D} = \mathcal{T}( \mathrm{f}^n_{2D}),\quad \mathrm{f}^l_{3D} = \mathcal{T}( \mathrm{f}^n_{3D}),
\end{equation}
where $\mathcal{T}(\cdot)$ is the shared temporal layers, $f^{n}_{2D} \in \mathbb{R}^{L \times C}$.

\subsubsection{2D/3D decoder.}
As for the 2D decoder, we use a single transformer block including the self-attention layer and the feed forward networks (FFNs), aiming to encourage the shared transformer layers to learn better universal features. 

The 3D decoder shares the same architecture with the 2D decoder in \stage{1}. However, we found that it is difficult to regress 3D human pose through a light decoder without performance drop in \stage{2}. Therefore, we additionally use stride-transformer layers~\cite{li2022exploiting} to enhance the lifting ability of the 3D decoder on 3D HPE task. This process can be formulated as:
\begin{equation}
    \mathrm{P}^{out}_{2D} = \mathcal{D}_{2D}(\mathrm{f}^l_{2D}),\quad  \mathrm{P}^{out}_{3D}= \mathcal{D}_{3D}(\mathrm{f}^l_{3D}),
\end{equation}
where $\mathrm{P}^{out}_{2D}$ and $\mathrm{P}^{out}_{3D}$ are the output 2D and 3D pose sequence, $\mathcal{D}_{2D}(\cdot)$ and $\mathcal{D}_{3D}(\cdot)$ are 2D and 3D encoders, and $\mathrm{f}^l_{2D}$ and $\mathrm{f}^l_{3D}$ are 2D and 3D pose embedding after shared transformer layers which share the common space.

\subsection{\Stage{1}: Pre-training via Masked Pose Modeling}
\subsubsection{Masked sampling strategies}
We conduct randomly masked sampling \textit{frame-by-frame} as shown in Figure~\ref{fig:mask}. In this case, the masked joints between adjacent frames could not be exactly the same. The masked joints are replaced by a shared learnable vector $v^S$ before the encoder. It is also padded by a shared learnable vector $v^T$ before the decoder, which is similar to~\cite{he2022masked}. 
% We claim this kind of masked sampling strategy can help to learn both spatial and temporal relations, as evidenced from ablation studies among different masked sampling strategies in Section~\ref{mask}. 

\begin{figure}[h]
    \centering
    \vspace{-10pt}
    \includegraphics[width=0.5\linewidth]{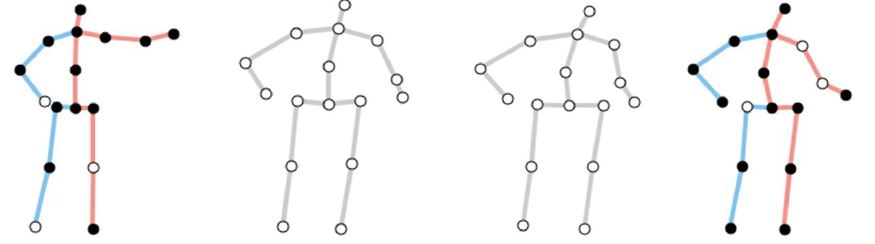}
    \caption{Illustration of spatial-temporal masking sampling strategy.}
    \label{fig:mask}
    \vspace{-30pt}
\end{figure}

\subsubsection{Masked 2D/3D pose modeling.}
We use the L2 loss between the masked and reconstructed 2D/3D pose sequences: 
\begin{equation}
\mathrm{P}^{in}_{i} = \mathrm{M}_{i},\quad \mathrm{P}^{out}_{i} = (\mathcal{D}_{i} \circ \mathcal{T} \circ \mathcal{S} \circ \mathcal{E}_{i})(\mathrm{M}_{i}),
\end{equation}
for $i \in \{2D, 3D\}$ where $\mathrm{M}_{i}$ is the masked pose sequence. Then the masked pose modeling loss can be calculated by:

\begin{equation}
\mathcal{L}_{\text{m2m}} = \lVert  \mathrm{P}^{out}_{2D}-\mathrm{P}^{in}_{2D} \rVert_2, \quad\mathcal{L}_{m3m} = \lVert  \mathrm{P}^{out}_{3D}-\mathrm{P}^{in}_{3D} \rVert_2.
\label{eq: m3m}
\end{equation}
We conduct those two pre-training tasks iteratively with the same frequency. The overall loss function in \stage{1} is:
\begin{equation}
    \mathcal{L}_{\text{\stage{1}}} = \lambda_{\text{m2m}} \mathcal{L}_{\text{m2m}}+\lambda_{\text{m3m}}\mathcal{L}_{\text{m3m}},
\end{equation}
where $\lambda_{\text{m2m}}$, $\lambda_{\text{m3m}}$ are weight factors.

\subsection{\Stage{2}: Fine-tuning via Full-supervision}
\subsubsection{3D HPE.}
In \stage{2}, we use 2D-3D pose pairs to further perform fine-tuning. We input a 2D sequence to the model and get the 3D pose of the middle frame following~\cite{pavllo20193d} as
\begin{equation}
\mathrm{P}^{out}_{3D} = ( \mathcal{D}_{3D} \circ \mathcal{T} \circ \mathcal{S} \circ \mathcal{E}_{2D} )(\mathrm{P}_{2D}).
\end{equation}

Here we still use L2 loss between the prediction and ground truth 3D pose formulated as: 
\begin{equation}
      \mathcal{L}_{\text{ft}} = \lVert \mathrm{PM}_{3D}^{out} - \mathrm{PM}_{3D}^{gt} \rVert_2,
      \label{eq:ft}
\end{equation}
where $\mathrm{PM}_{3D}^{out}$ and $\mathrm{PM}_{3D}^{gt}$ represent the single pose in the middle frame of the pose sequence and the corresponding ground truth, respectively.

We also utilize a simple linear projection after the shared layer to obtain the whole 3D pose sequence as extra supervision. The loss function denotes as:
\begin{equation}
      \mathcal{L}_{\text{ft-seq}} = \lVert \mathrm{P}^{3D}_{out} - \mathrm{P}^{3D}_{gt} \rVert_2.
\end{equation}

The overall loss function of 3D HPE in \stage{2} is:
\begin{equation}
    \mathcal{L}_{\text{3dhpe}} = \lambda_{\text{ft}} \mathcal{L}_{\text{ft}}+\lambda_{\text{ft-seq}}\mathcal{L}_{\text{ft-seq}},
\end{equation}
where $\lambda_{\text{ft}}$ and $\lambda_{\text{ft-seq}}$ are weight factors.

\subsubsection{Pose completion.} For partial 3D to 3D pose completion task, the loss is the same as Equation~\eqref{eq: m3m} except that the masking strategy is set to tube masking, in which temporal mask ratio is set to 0 and masked joints between adjacent frames are the same. For the partial 2D to 3D pose completion task, we only use L2 loss same as Equation~\eqref{eq:ft} with incomplete 2D pose as input.

\section{Experiments}
\subsection{Datasets and Metrics}
We use three widely used 2D-3D human pose datasets to train our network including Human3.6M~\cite{ionescu2013human3}, MPI-INF-3DHP~\cite{mehta2017monocular}, and AMASS~\cite{mahmood2019amass}. We evaluate the performance on Human3.6M and MPI-INF-3DHP.

\subsubsection{Human3.6M.} Human3.6M dataset, which contains 3.6 million frames of corresponding 2D and 3D human poses, is a video and mocap dataset of 5 female and 6 male subjects. According to the previous works~\cite{drover2018can,fang2018learning,chen2021anatomy}, we choose 5 subjects (S1, S5, S6, S7, S8) for training, and the other 2 subjects (S9 and S11) for evaluation. We report the Mean Per Joint Position Error (MPJPE) as the metric of Protocol \#1 as well as Procrusts analysis MPJPE (P-MPJPE) as the metric of Protocol \#2.

\subsubsection{MPI-INF-3DHP.}
Compared to Human3.6M, MPI-INF-3DHP is a more challenging 3D human pose dataset captured in the wild. There are 8 subjects with 8 actions captured by 14 cameras covering a greater diversity of poses. We follow the setting of evaluation in~\cite{mehta2017monocular} with three common metrics PCK, AUC, and MPJPE.
% \vspace{-3pt}
\subsubsection{AMASS.}
AMASS~\cite{mahmood2019amass} is a large database of human motion unifying different optical marker-based motion capture datasets. The consistent representation of AMASS makes it readily useful for animation, visualization, and generating training data for deep learning. It totally has more than 40 hours of duration over 300 subjects, and more than 11,000 motions are involved.

\vspace{-5pt}
\subsection{Implementation details}
\subsubsection{Training settings.} We implement our proposed method using PyTorch~\cite{paszke2019pytorch} on 4 NVIDIA RTX 3090 GPUs. For training \stage{1}, we use AdamW~\cite{loshchilov2017decoupled} optimizer and train for 15 epochs. In \stage{2}, we fine-tune our model for 40 epochs on Human3.6M or 100 epochs on MPI-INF-3DHP. The learning rate is set to $1e^{-4}$ for \stage{1} and $1.6e^{-3}$ for \stage{2}, and we set a weight decay factor of 0.97 for each epoch. As a video-based method, we choose two different input sequence lengths of 81 and 243. The loss weight factors $\lambda_{\text{m2m}}$, $\lambda_{\text{m3m}}$, $\lambda_{\text{ft}}$ and $\lambda_{\text{ft-seq}}$ are set to 2.0, 3.0, 1.0, 1.0, respectively.  We conduct experiments under mask ratio $r_s^* = 5/17 $ in spatial and $r_t^* = 60\%$ in temporal, which leads to the masking ratio of $71.8\%$ in total. The detailed ablation experiment of mask ratio is in the Appendix.
We additionally use a refine module as \cite{cai2019exploiting} to enhance the performance on the Human3.6M benchmark. Note that we use all three datasets  (\textbf{18.53M} samples in total) for the pertaining in Table~\ref{tab:3dhp} 
 and exclude 3DHP (\textbf{13.31M} samples in total) in pre-training of other experiments on Human3.6M due to different keypoint definitions.

\begin{table*}[t]
\scriptsize

\centering
  \caption{Quantitative comparison of Mean Per Joint Position Error between the estimated 3D pose and the ground truth 3D pose on Human3.6M under Protocols \#1 \& \#2 using the detected 2D pose as input. Top-table: results under Protocol \#1 (MPJPE). Bottom-table: results under Protocol \#2 (P-MPJPE). $\dagger$ denotes a Transformer-based model. The best and second scores are marked in bold and underline.}

\resizebox{\linewidth}{!}{
\begin{tabular}{lcccccccccccccccccc}

\toprule

\multicolumn{1}{c|}{\textbf{Protocol \#1 (GT)}}       & \multicolumn{1}{c|}{Venue} & \multicolumn{1}{c|}{\# Frames}   & Dir. & Disc. & Eat. & Greet & Phone & Photo & Pose & Purch. & Sit  & SitD. & Somke & Wait & WalkD. & Walk & \multicolumn{1}{c|}{WalkT.} & Average \\ 
\midrule

\multicolumn{1}{l|}{Lin \etal \cite{lin2019trajectory}}    & \multicolumn{1}{c|}{ BMVC'18} &  \multicolumn{1}{c|}{50}   & -  & -  & - & -  & -  & -  & - & -   & - & -  & -  & - & -   & - & \multicolumn{1}{c|}{-}   & 32.8   \\
\multicolumn{1}{l|}{Liu \etal \cite{liu2020attention}}   & \multicolumn{1}{c|}{CVPR'20}  &  \multicolumn{1}{c|}{243} & 34.5 & 37.1  & 33.6 & 34.2  & 32.9  & 37.1  & 39.6 & 35.8   & 40.7 &41.4 &33.0 &33.8 &33.0& 26.6   & \multicolumn{1}{c|}{26.9}   & 34.7    \\

\multicolumn{1}{l|}{Zeng \etal \cite{zeng2020srnet}}    & \multicolumn{1}{c|}{ECCV'20}  & \multicolumn{1}{c|}{243}  & 34.8 & 32.1 & 28.5 & 30.7  & 31.4 & 36.9 & 35.6 & 30.5 & 38.9&  40.5 &32.5 &31.0 &29.9& 22.5 & \multicolumn{1}{c|}{24.5}   &   32.0    \\

\multicolumn{1}{l|}{Zheng \etal \cite{zheng20213d}$\dagger$}   & \multicolumn{1}{c|}{ICCV'21} &\multicolumn{1}{c|}{81}  & 30.0 &33.6 &29.9 &31.0 &30.2 &33.3 &34.8 &31.4 &37.8 &38.6 &31.7 &31.5 &29.0 &23.3  & \multicolumn{1}{c|}{23.1}   &  31.3    \\  		

\multicolumn{1}{l|}{Shan \etal \cite{shan2021improving}}   & \multicolumn{1}{c|}{MM'21} & \multicolumn{1}{c|}{243}  & 29.5 &30.8 &28.8 &29.1 &30.7 &35.2 &31.7& 27.8 &34.5 &36.0 &30.3 & 29.4 &28.9 &24.1  & \multicolumn{1}{c|}{24.7}   & 30.1
\\

\multicolumn{1}{l|}{Shan \etal \cite{shan2022p}$\dagger$}  & \multicolumn{1}{c|}{ECCV'22 }& \multicolumn{1}{c|}{81} & 28.5& 30.1 &28.6& 27.9 & 29.8& 33.2 & 31.3 & 27.8& 36.0& 37.4& 29.7&29.5& 28.1 &21.0   & \multicolumn{1}{c|}{21.0}   & 29.3  \\ 

\multicolumn{1}{l|}{Zhang \etal \cite{Zhang2022MIXSTE}$\dagger$}        & \multicolumn{1}{c|}{CVPR'22}& \multicolumn{1}{c|}{243} & \textbf{21.6} & \textbf{22.0}  & \textbf{20.4 } & \textbf{21.0} & \textbf{20.8}  & \textbf{24.3}  & \textbf{24.7}  & 21.9 & \textbf{26.9} & \textbf{24.9} & \textbf{ 21.2} & \textbf{21.5}  & \textbf{20.8} &\textbf{14.7} & \multicolumn{1}{c|}{\textbf{15.7}}   & \textbf{21.6}    \\

\midrule
\rowcolor[gray]{0.9}
\multicolumn{1}{l|}{\mpm~(\textit{ours})$\dagger$} & \multicolumn{1}{c|}{$-$}  & \multicolumn{1}{c|}{81}   & 30.4 & 32.2 & 29.2 & 28.4 & 29.3 & 33.7 & 34.2  & 29.3 & 40.0 & 35.6  & 30.8 & 30.7  & 27.4& 20.7 & \multicolumn{1}{c|}{20.7}   &  30.2 \\ 

\rowcolor[gray]{0.9}
\multicolumn{1}{l|}{\mpm~(\textit{ours})$\dagger$} & \multicolumn{1}{c|}{$-$} & \multicolumn{1}{c|}{243}   & 26.7 & \underline{28.1} & \underline{27.1} & \underline{26.2} & \underline{27.8} & \underline{30.9}  & 30.5  & 26.0   &\underline{33.9} & \underline{33.1}  & \underline{27.3}  & 28.1 & \underline{24.7}  & \underline{18.3} & \multicolumn{1}{c|}{\underline{18.5}}   &  \underline{27.2} \\ 
\midrule

\midrule
\multicolumn{1}{c|}{\textbf{Protocol \#1 (CPN)}}  &  \multicolumn{1}{c|}{Venue} & \multicolumn{1}{c|}{\# Frames}  & Dir. & Disc. & Eat. & Greet & Phone & Photo & Pose & Purch. & Sit  & SitD. & Somke & Wait & WalkD. & Walk & \multicolumn{1}{c|}{WalkT.} & Average \\ 
\midrule

\multicolumn{1}{l|}{Pavllo \etal \cite{pavllo20193d}} & \multicolumn{1}{c|}{CVPR'19 } & \multicolumn{1}{c|}{243} & 45.2 & 46.7  & 43.3 & 45.6  & 48.1  & 55.1  & 44.6 & 44.3   & 57.3 & 65.8  & 47.1  & 44.0 & 49.0   & 32.8 &
\multicolumn{1}{c|}{33.9}   & 46.8    \\

\multicolumn{1}{l|}{Lin \etal \cite{lin2019trajectory}}     & \multicolumn{1}{c|}{BMVC'19} & \multicolumn{1}{c|}{243}    & 42.5 & 44.8  & 42.6 & 44.2  & 48.5  & 57.1  & 52.6 & 41.4   & 56.5 & 64.5  & 47.4  & 43.0 & 48.1   & 33.0 & \multicolumn{1}{c|}{35.1}   & 46.6    \\

\multicolumn{1}{l|}{Liu \etal \cite{liu2020attention}}   & \multicolumn{1}{c|}{CVPR'20} & \multicolumn{1}{c|}{243} &   41.8 & 44.8  & 41.1 & 44.9  & 47.4  & 54.1  & 43.4 & 42.2   & 56.2 & 63.6  & 45.3  & 43.5 & 45.3   & 31.3 & \multicolumn{1}{c|}{32.2}   & 45.1    \\
\multicolumn{1}{l|}{Zeng \etal \cite{zeng2020srnet}}       & \multicolumn{1}{c|}{ECCV'20}& \multicolumn{1}{c|}{243}    & 46.6 & 47.1  & 43.9 & 41.6  &45.8  & 49.6  & 46.5 & 40.0   & 53.4 & 61.1  & 46.1  & 42.6 &43.1   & 31.5 & \multicolumn{1}{c|}{32.6}   & 44.8    \\

\multicolumn{1}{l|}{Chen \etal \cite{chen2021anatomy}}  & \multicolumn{1}{c|}{TCSVT'21} & \multicolumn{1}{c|}{81} & 42.1 & 43.8  &  41.0 & 43.8  & 46.1  & 53.5  & 42.4 & 43.1   & 53.9 & 60.5  & 45.7  & 42.1 & 46.2   & 32.2 & \multicolumn{1}{c|}{33.8}   & 44.6    \\

\multicolumn{1}{l|}{Zheng \etal \cite{zheng20213d}$\dagger$}         &  \multicolumn{1}{c|}{ICCV'21}& \multicolumn{1}{c|}{81}   &  41.5 & 44.8  & 39.8 & 42.5  & 46.5  &51.6  & 42.1 & 42.0   & \underline{53.3} & 60.7  & 45.5  & 43.3 & 46.1   & 31.8 & \multicolumn{1}{c|}{32.2}   & 44.3   \\ 

\multicolumn{1}{l|}{Chen \etal \cite{chen2021anatomy}}         &  \multicolumn{1}{c|}{TCSVT'21} & \multicolumn{1}{c|}{243}  &  41.4 & 43.5  & 40.1  & 42.9  & 46.6  &51.9  & 41.7 & 42.3   &53.9 & 60.2  & 45.4  & 41.7 & 46.0   & 31.5 & \multicolumn{1}{c|}{32.7}   & 44.1   \\ 

\multicolumn{1}{l|}{Hu \etal \cite{hu2021conditional}}         &  \multicolumn{1}{c|}{MM'21} & \multicolumn{1}{c|}{243}  &  \underline{38.0} & 43.3  & \underline{39.1}  & \textbf{39.4}  & 45.8  &53.6  & 41.4 & 41.4   &55.5 & 61.9  & 44.6  & 41.9 & 44.5   & 31.6 & \multicolumn{1}{c|}{29.4}   & 43.4   \\ 

\multicolumn{1}{l|}{Li \etal \cite{li2022exploiting}$\dagger$}         &  \multicolumn{1}{c|}{TMM'22} & \multicolumn{1}{c|}{351}   &  39.9 & 43.4  & 40.0  & 40.9  & 46.4  & 50.6  & 42.1 & 39.8   &55.8 & 61.6  & 44.9  & 43.3 & 44.9   & 29.9 & \multicolumn{1}{c|}{30.3}   & 43.6   \\ 

\multicolumn{1}{l|}{Shan \etal \cite{shan2022p}$\dagger$}         & \multicolumn{1}{c|}{ECCV'22} &\multicolumn{1}{c|}{243}   &  38.4 & 42.1  & 39.8 & 40.2  & 45.2 & \textbf{48.9}  & 40.4 & \textbf{38.3}   &53.8 & \underline{57.3}  & 43.9  & 41.6 & 42.2   & 29.3 & \multicolumn{1}{c|}{29.3}   & \underline{42.1}   \\ 

\multicolumn{1}{l|}{Zhang \etal \cite{Zhang2022MIXSTE}$\dagger$}        & \multicolumn{1}{c|}{CVPR'22}& \multicolumn{1}{c|}{243} & \textbf{37.6} & \textbf{40.9}  & \textbf{37.3} & \underline{39.7} & \textbf{42.3}  & \underline{49.9}  & \underline{40.1}  & 39.8 & \textbf{51.7} & \textbf{55.0} & \textbf{42.1} & \textbf{39.8}  & \textbf{41.0} &\textbf{27.9 } & \multicolumn{1}{c|}{\textbf{27.9}}   & \textbf{40.9}    \\

\midrule
\rowcolor[gray]{0.9}
\multicolumn{1}{l|}{\mpm~(\textit{ours})$\dagger$}  & \multicolumn{1}{c|}{$-$} & \multicolumn{1}{c|}{81}   & 39.7   & 43.2  & 39.9 & 42.7  & 45.8  & 52.7 & 41.7 &  41.1 & 55.3 & 60.5  & 45.1  & 42.7 & 43.8 & 29.4 & \multicolumn{1}{c|}{30.2}   &  43.6 \\

\rowcolor[gray]{0.9}
\multicolumn{1}{l|}{\mpm~(\textit{ours})$\dagger$}  & \multicolumn{1}{c|}{$-$} & \multicolumn{1}{c|}{243} &  39.7 & \underline{42.1}  & 39.2  & 40.3  & \underline{45.2}  &50.9  & \textbf{40.1} & \underline{39.6}   &53.9 & 59.4 & \underline{43.0}  &\underline{41.3} & \underline{42.2}   & \underline{28.9} & \multicolumn{1}{c|}{\underline{29.1}}   & 42.3   \\ 
\midrule

\midrule
\multicolumn{1}{c|}{\textbf{Protocol \#2 (CPN)}} & \multicolumn{1}{c|}{Venue}  &  \multicolumn{1}{c|}{\# Frames}      & Dir. & Disc. & Eat. & Greet & Phone & Photo & Pose & Purch. & Sit  & SitD. & Somke & Wait & WalkD. & Walk & \multicolumn{1}{c|}{WalkT.} & Average \\ 
\midrule

\multicolumn{1}{l|}{Lin \etal \cite{lin2019trajectory} }    & \multicolumn{1}{c|}{BMVC'19}  & \multicolumn{1}{c|}{50}  & 32.5 & 35.3  & 34.3 & 36.2  & 37.8  & 43.0  & 33.0 & 32.2   & 45.7 & 51.8  & 38.4  & 32.8 & 37.5   & 25.8 & \multicolumn{1}{c|}{28.9}   & 36.8    \\
\multicolumn{1}{l|}{Pavllo \etal \cite{pavllo20193d} }  & \multicolumn{1}{c|}{CVPR'19}& \multicolumn{1}{c|}{243} & 34.1 & 36.1  & 34.4 & 37.2  & 36.4  & 42.2  & 34.4 & 33.6   & 45.0 & 52.5  & 37.4  & 33.8 & 37.8   & 25.6 & \multicolumn{1}{c|}{27.3}   & 36.5    \\

\multicolumn{1}{l|}{Xu \etal \cite{xu2020deep}}    & \multicolumn{1}{c|}{CVPR'20}  & \multicolumn{1}{c|}{50}  & \underline{31.0} & 34.8  & 34.7 & 34.4  & 36.2  & 43.9  & 31.6 & 33.5   & \textbf{42.3} & 49.0  & 37.1  & 33.0 & 39.1   & 26.9 & \multicolumn{1}{c|}{31.9}   & 36.2    \\

\multicolumn{1}{l|}{Liu \etal \cite{liu2020attention}}   & \multicolumn{1}{c|}{CVPR'20}& \multicolumn{1}{c|}{243}  &32.3 &  35.2  & 33.3 & 35.8  &35.9  & 41.5  & 33.2 & 32.7   & 44.6 & 50.9  & 37.0  & 32.4 & 37.0   & 25.2 & \multicolumn{1}{c|}{27.2}   & 35.6    \\

\multicolumn{1}{l|}{Chen \etal \cite{chen2021anatomy}}  & \multicolumn{1}{c|}{TCSVT'21 }& \multicolumn{1}{c|}{81} & 33.1 & 35.3  & 33.4 & 35.9  & 36.1  & 41.7  & 32.8 & 33.3   & 42.6 & 49.4  & 37.0  &32.7 & 36.5   & 25.5 & \multicolumn{1}{c|}{27.9}   & 35.6    \\ 

\multicolumn{1}{l|}{Zheng \etal \cite{zheng20213d}$\dagger$}        & \multicolumn{1}{c|}{ICCV'21} & \multicolumn{1}{c|}{81} &32.5 &34.8  & 32.6 & 34.6  & 35.3  & 39.5  & 32.1 & 32.0   & 42.8 & 48.5  & \underline{34.8}  & 32.4 &35.3   & 24.5 & \multicolumn{1}{c|}{26.0}   & 34.6   \\ 

\multicolumn{1}{l|}{Shan \etal \cite{shan2022p}$\dagger$}        & \multicolumn{1}{c|}{ECCV'22}& \multicolumn{1}{c|}{243} & 31.3 & 35.2  & 32.9 & 33.9 & 35.4  & \textbf{39.3}  & 32.5  & 31.5 & 44.6 & \underline{48.2} & 36.3 & 32.9  & 34.4  &23.8 & \multicolumn{1}{c|}{\underline{23.9}}   & 34.4    \\

\multicolumn{1}{l|}{Zhang \etal \cite{Zhang2022MIXSTE}$\dagger$}        & \multicolumn{1}{c|}{CVPR'22}& \multicolumn{1}{c|}{243} & \textbf{30.8} & \textbf{33.1}  & \textbf{30.3} & \textbf{31.8} & \textbf{33.1}  & \underline{39.1}  & \textbf{31.1}  & \textbf{30.5} & \underline{42.5} & \textbf{44.5} & \textbf{34.0} & \textbf{30.8}  & \textbf{32.7} &\textbf{22.1} & \multicolumn{1}{c|}{\textbf{22.9}}   & \textbf{32.6}    \\

\midrule
\rowcolor[gray]{0.9}
\multicolumn{1}{l|}{\mpm~(\textit{ours})$\dagger$}  & \multicolumn{1}{c|}{$-$} & \multicolumn{1}{c|}{81}   & 31.9   & 35.3  & 33.2 & 35.0  & 35.6  & 40.5 & 33.0 &  32.4 & 44.8 & 48.7  & 37.3 & 33.4 & 34.4 & 24.2 & \multicolumn{1}{c|}{25.7}   &  35.0   \\ 

\rowcolor[gray]{0.9}
\multicolumn{1}{l|}{\mpm~(\textit{ours})$\dagger$}  & \multicolumn{1}{c|}{$-$} & \multicolumn{1}{c|}{243}   & 32.4 & \underline{34.7} & \underline{32.3} & \underline{33.6} & \underline{35.3}  & 40.4 & 31.8 & \underline{31.5}  & 44.3 & 49.4  & 35.4 & \underline{32.2} & \underline{34.2} & \underline{23.7} & \multicolumn{1}{c|}{24.3}   & \underline{34.4}  \\ 
\bottomrule

\end{tabular}
}

\label{tab: h36m}
\end{table*}

\begin{table}[h]
% \tiny
\scriptsize
\centering
\caption{Quantitative comparison with previous methods on MPI-INF-3DHP. PCK, AUC, and MPJPE metrics are reported. The best and second scores are marked in bold and underlined respectively.}
\resizebox{0.6\linewidth}{!}
{
\begin{tabular}{l|c|c|ccc}
\toprule
Method & Venue & \# Frames & PCK~($\uparrow$) & AUC~($\uparrow$) & MPJPE~($\downarrow$) \\
\midrule

Pavllo \etal \cite{pavllo20193d} & CVPR'19 & 243 & 85.5 & 51.5 & 84.8  \\
Lin \etal \cite{lin2019trajectory} & BMVC'19  & 25 & 83.6 & 51.4 & 79.8  \\ 
Li \etal \cite{Li2020cascade3d} & CVPR'20   & 1 & 81.2 & 46.1 & 99.7  \\
Chen \etal \cite{chen2021anatomy}  & TCSVT'21 & 81  & 87.9 & 54.0 & 78.8  \\
Zheng \etal \cite{zheng20213d}     & ICCV'21  & 81   & 88.6  & 56.4 & 77.1  \\ 
Zhang \etal \cite{Zhang2022MIXSTE} & CVPR'22 & 243 & 94.4 & 66.5 & 54.9\\
Hu \etal \cite{hu2021conditional} & MM'21 & 81 & 97.9 & 69.5 & 42.5 \\
Shan \etal \cite{shan2022p} & ECCV'22 & 81 & 97.9& 75.8 & 32.2 \\
\midrule

\rowcolor[gray]{0.9}
\mpm~(\textit{ours}) & $-$ & 81 & \textbf{98.6}  & \underline{79.5} & \underline{29.8}\\
\rowcolor[gray]{0.9}
\mpm~(\textit{ours}) & $-$ & 243 & \underline{98.4} & \textbf{80.0} & \textbf{29.1} \\
\bottomrule
\end{tabular}
}
\label{tab:3dhp}
\end{table}

\vspace{-5pt}
\subsection{Comparison with State-of-the-Art Methods}
\subsubsection{3D Human Pose Estimation Task}

\paragraph{\textbf{Results on Human3.6M.}}
We compare our methods with existing state-of-the-art methods on Human3.6M dataset. We report the results of all 15 actions in the test subjects (S9 and S11) as shown in Table~\ref{tab: h36m}. Following~\cite{li2022exploiting,zheng20213d,shan2022p}, we use 2D poses detected by the CPN~\cite{chen2018cascaded} detector. Our method achieves comparable performance. Notice that the FLOPs of \cite{Zhang2022MIXSTE}
is \textbf{156,735M} while \textbf{MPM}’s is only \textbf{2387M} with a relatively small difference on performance.

\paragraph{\textbf{Results on MPI-INF-3DHP.}}
We report our result and compare it with state-of-the-art methods on MPI-INF-3DHP\cite{mykhaylo20143dhp} dataset as shown in Table \ref{tab:3dhp}. We only adopt the ground truth of 2D poses as input.

\begin{figure*}[!t]
    \centering
    \includegraphics[width=0.97\linewidth]{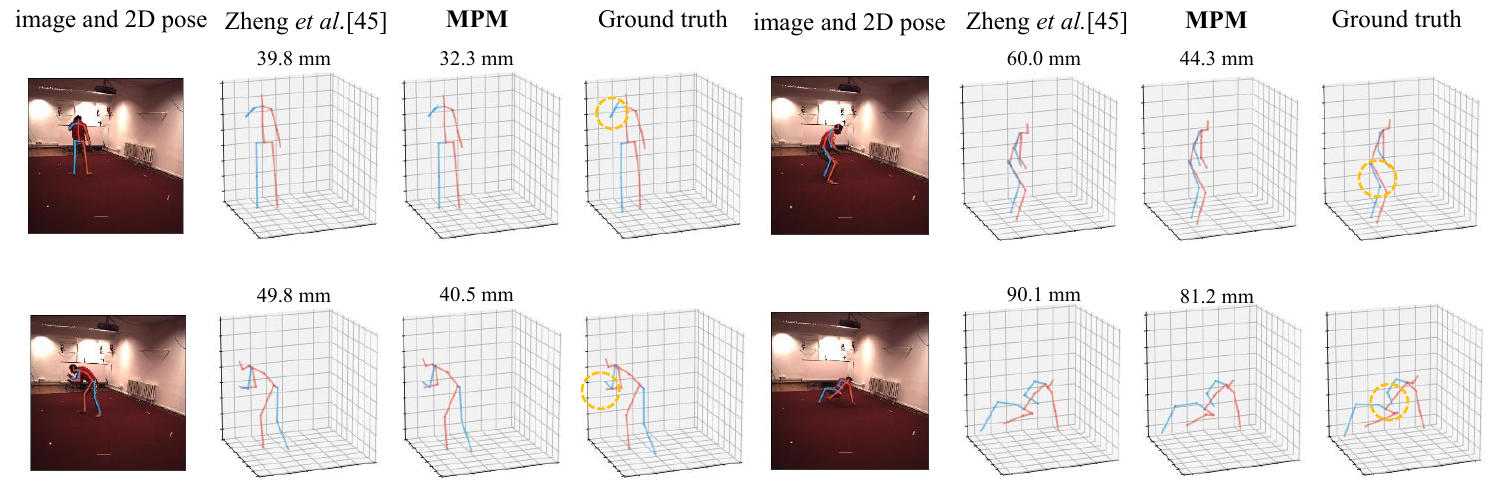}
    \caption{Qualitative visualization comparison with previous method~\cite{zheng20213d} and ground truth on Human3.6M benchmarks.}
    \label{fig:vis}
\end{figure*}

\paragraph{\textbf{Qualitative visualization.}}
Figure~\ref{fig:vis} demonstrates the qualitative visualization comparison with the previous works on Human3.6M  benchmark.

\begin{table}[!h]
\caption{Recover 3D pose from partial 2D observation under Protocol \#1. $S$ denotes sample number in multi-hypothesis setting.}
\label{tab:c2d}
\centering{
\resizebox{0.7\linewidth}{!}{
\begin{tabular}{c|cccc}
\toprule
Occ. Body Parts & \mpm~(\textit{ours}) & GFPose~$(S=1)$~\cite{ci2023gfpose}& GFPose~$(S=200)$ & Li~\etal~\cite{li2019generating} \\
\midrule
1 Joint & \textbf{40.5 } & 71.7 & 37.8 & 58.8 \\
2 Joints & \textbf{41.5} & 78.3 & 39.6 & 64.6 \\
\midrule
2 Legs & \textbf{62.1} & 108.6 & 53.5 & $-$ \\
2 Arms & \textbf{77.0} & 116.9 & 60.0 & $-$ \\
Left Leg + Left Arm & \textbf{56.9} & 106.8& 54.6 & $-$ \\
Right Leg + Right Arm & \textbf{56.3} & 109.7 & 53.1 & $-$ \\
\bottomrule
\end{tabular}
}
}
\end{table}

\begin{table}[!h]
\caption{Recover 3D pose from partial 3D observation under Protocol \#1. $S$ denotes sample number in multi-hypothesis setting.}
\label{tab:c3d}
\centering{
\resizebox{0.65\linewidth}{!}{
\begin{tabular}{c|ccc}
\toprule
Occ. Body Parts & \mpm~(\textit{ours}) & GFPose~$(S=1)$~\cite{ci2023gfpose} & GFPose~$(S=200)$ \\
\midrule
Right Leg & \textbf{13.0} & 13.0 & 5.2 \\
Left Leg & \textbf{12.4}& 14.3 & 5.8 \\
Left Arm & \textbf{24.0} &  25.5& 9.4 \\
Right Arm & \textbf{21.6} &  22.4 & 8.9\\
\bottomrule
\end{tabular}
}
}
\end{table}

\subsubsection{3D Pose Completion Task}
\paragraph{\textbf{From incomplete 2D pose.}}
In real-world applications, 2D pose estimation algorithms and MoCap systems often suffer from occlusions, which result in incomplete detected 2D poses. Fine-tuned by masked 2D pose lifting, \mpm~can also help to recover an intact 3D human pose from incomplete 2D observations at either the joint level or body part level. As shown in Table~\ref{tab:c2d}, our method achieves comparable results even though GFPose~\cite{ci2023gfpose} uses 200 samples in multi-hypothesis setting and also with camera intrinsic, Li~\etal~\cite{li2019generating} train different models to deal with varying numbers of missing joints.

\paragraph{\textbf{From incomplete 3D pose.}}
Fitting to partial 3D observations also has many potential downstream applications, \eg, generating the missing part of the body for a person in VR. Shown in Table~\ref{tab:c3d}, \mpm~can be directly used or after fine-tuning to recover missing 3D body parts given partial 3D observations.

\subsection{Ablation Study}
In this section, we conduct ablation studies to answer the following questions. More ablation studies can be found in the \textbf{supplementary materials} due to the page limitation.

\vspace{-10pt}
\subsubsection{Is Unifying 2D and 3D Representation Beneficial to 3D HPE?}
To prove the advantage of unifying 2D and 3D representation, we conduct experiments for the following 3 settings: No pre-train model and fine-tuning, only using 2D information to pre-train and fine-tuning, unifying 2D and 3D representation, and fine-tuning. Table~\ref{tab:pretext} demonstrates that using both 2D and 3D information in \stage{1} brought the best performance in the 3D HPE task. 
Table~\ref{tab:param} demonstrates the number of parameters in different components, and it shows that shared parameters take the majority. This is also a piece of evidence that 2D and 3D poses can share a common latent feature space.
\vspace{-10pt}
\begin{table}[!h]
\centering
\caption{Analysis on computational complexity of each module.}
\resizebox{0.5\linewidth}{!}
{
\begin{tabular}{c|cc}
\toprule
Module & \# Params. [K] & FLOPs [M] \\
\midrule
2D Encoder & 17.9 & 8.9\\ 
3D Encoder & 26.1 & 12.9 \\ 
Shared layer & 3289.1 & 1796.3\\
2D Decoder & 532.2 & 259.3\\
3D Decoder (\stage{1}) & 534.3 & 261.3 \\
3D Decoder (\stage{2}) & 3941.1 & 380.6\\
\bottomrule
\end{tabular}
}
\label{tab:param}
\end{table}
\vspace{-30pt}
\begin{table}[!h]
    \centering
    \caption{Performance under different pre-text task combinations on Human3.6M benchmark. }
    \resizebox{0.6\linewidth}{!}{
    \begin{tabular}{cc|c|cc}
    \toprule
    2D comp. & 3D comp.   & \# Frames &  MPJPE~($\downarrow$) & P-MPJPE~($\downarrow$) \\ 
    \midrule
    \ding{56} & \ding{56}   & 243 &  43.1  &  34.9         \\
    \ding{52} & \ding{56}   & 243 &  42.9  &   34.6        \\
    \ding{52} & \ding{52}   & 243 & \textbf{42.3} & \textbf{34.4} \\
    \bottomrule
    \end{tabular}
    }
    \label{tab:pretext}
\end{table}

\vspace{-20pt}

\vspace{-10pt}
\begin{figure}[!h]
    \centering
    \includegraphics[width=0.55\linewidth]{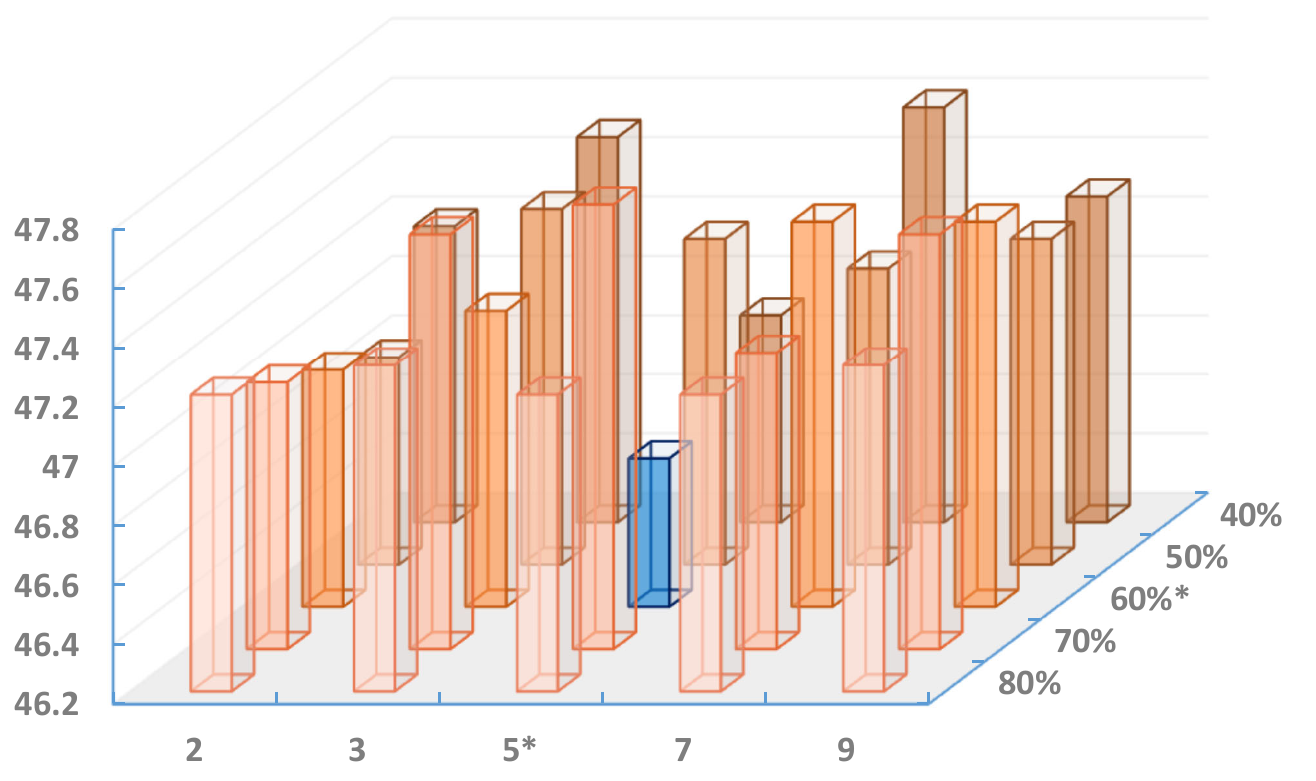}
    \caption{Grid search of masking number of joints $r_s$ (x-axis) and ratio $r_t$ (y-axis) for spatiotemporal masking. We report the performance over MPJPE (z-axis) on Human3.6M with 27 frames length. The optimal ratios are $r_s^* = 5/17 $ in spatial and $r_t^* = 60\%$ in temporal leads to $46.7$ mm of MPJPE. Detailed results are listed in the Appendix.}
    \label{fig:ratio}
\end{figure}
\vspace{-20pt}

\vspace{-10pt}
\subsubsection{What is the Best Masking Ratio in \mpm?}
Figure~\ref{fig:ratio} shows the influence of the masking ratio in a grid search in 27 frames setting. We claim that the masking ratio in spatial and temporal should be considered separately. The optimal ratios are $r_s^* = 5/17 $ in spatial and $r_t^* = 60\%$ in temporal, which leads to the masking ratio of $71.8\%$ in total. The optimal ratio is larger than $15\%$ of BERT~\cite{devlin2018bert} in NLP, similar to $75\%$ of MAE~\cite{he2022masked} in images, smaller than $95\%$ of VideoMAE~\cite{feichtenhofer2022masked} in videos and $90\%$ of P-STMO~\cite{shan2022p} in 2D pose sequences. The adjacent poses in a sequence are usually very similar, which is more redundant than words and images. 
Yet, modeling 2D and 3D poses together is more difficult than 2D poses alone. Therefore, the optimal ratio is smaller than that of P-STMO\cite{shan2022p}.

% \section{Limitation}

% Due to the computational resource limitation, we could not scale up our model into billion-size datasets. Although we use multiple datasets for training and evaluation, domain shift might happen due to poor diversity of training data.

\section{Conclusion and Limitation}

In conclusion, this paper presents \mpm, a novel framework for unifying 2D and 3D human pose representations in a shared feature space. While previous research has extensively explored estimating 3D human pose solely from 2D pose sequences, this work takes a significant step forward by integrating both modalities and leveraging a single-stream transformer-based architecture. We propose three pretext tasks: masked 2D pose modeling, masked 3D pose modeling, and masked 2D pose lifting. These tasks serve as pre-training objectives for their network, which is then fine-tuned using full-supervision. Notably, a high masking ratio of 71.8\% is employed with spatiotemporal mask sampling strategy. The \mpm~framework offers several advantages, as it enables handling multiple tasks within a single unified framework. Specifically, it facilitates 3D human pose estimation, 3D pose estimation from occluded 2D pose data, and 3D pose completion. The effectiveness of \mpm~is demonstrated through extensive experiments and ablation studies conducted on multiple widely used human pose datasets. Remarkably, \mpm~achieves state-of-the-art performance on MPI-INF-3DHP. Due to the computational resource limitation, we could not scale up our model into billion-size datasets. 
Moreover, contrastive paradigms like CLIP~\cite{radford2021learning} are also reasonable to help learning a unified representation of 2D and 3D human poses. The future research could also consider introducing RGB or depth information as other modality.

% \vspace{10pt}
\subsubsection{Acknowledgements.}
This work is supported by Zhejiang Provincial Natural Science Foundation of China under Grant No. LZ24F030005; The Fundamental Research Funds for the Central Universities (226-2024-00058); 

\bibliographystyle{splncs04}
\bibliography{ref}

\begin{thebibliography}{10}
\providecommand{\url}[1]{\texttt{#1}}
\providecommand{\urlprefix}{URL }
\providecommand{\doi}[1]{https://doi.org/#1}

\bibitem{mykhaylo20143dhp}
Andriluka, M., Pishchulin, L., Gehler, P., Schiele, B.: 2d human pose estimation: New benchmark and state of the art analysis. In: IEEE Conference on Computer Vision and Pattern Recognition (CVPR) (June 2014)

\bibitem{cai2019exploiting}
Cai, Y., Ge, L., Liu, J., Cai, J., Cham, T.J., Yuan, J., Thalmann, N.M.: Exploiting spatial-temporal relationships for 3d pose estimation via graph convolutional networks. In: Proceedings of the IEEE/CVF international conference on computer vision. pp. 2272--2281 (2019)

\bibitem{chai2023global}
Chai, W., Jiang, Z., Hwang, J.N., Wang, G.: Global adaptation meets local generalization: Unsupervised domain adaptation for 3d human pose estimation. arXiv preprint arXiv:2303.16456  (2023)

\bibitem{chen20173d}
Chen, C.H., Ramanan, D.: 3d human pose estimation= 2d pose estimation+ matching. In: Proceedings of the IEEE conference on computer vision and pattern recognition. pp. 7035--7043 (2017)

\bibitem{chen2021anatomy}
Chen, T., Fang, C., Shen, X., Zhu, Y., Chen, Z., Luo, J.: Anatomy-aware 3d human pose estimation with bone-based pose decomposition. IEEE Transactions on Circuits and Systems for Video Technology  \textbf{32}(1),  198--209 (2021)

\bibitem{chen2020uniter}
Chen, Y.C., Li, L., Yu, L., El~Kholy, A., Ahmed, F., Gan, Z., Cheng, Y., Liu, J.: Uniter: Universal image-text representation learning. In: Computer Vision--ECCV 2020: 16th European Conference, Glasgow, UK, August 23--28, 2020, Proceedings, Part XXX. pp. 104--120. Springer (2020)

\bibitem{chen2018cascaded}
Chen, Y., Wang, Z., Peng, Y., Zhang, Z., Yu, G., Sun, J.: Cascaded pyramid network for multi-person pose estimation. In: Proceedings of the IEEE conference on computer vision and pattern recognition. pp. 7103--7112 (2018)

\bibitem{ci2023gfpose}
Ci, H., Wu, M., Zhu, W., Ma, X., Dong, H., Zhong, F., Wang, Y.: Gfpose: Learning 3d human pose prior with gradient fields. In: Proceedings of the IEEE/CVF Conference on Computer Vision and Pattern Recognition. pp. 4800--4810 (2023)

\bibitem{devlin2018bert}
Devlin, J., Chang, M.W., Lee, K., Toutanova, K.: Bert: Pre-training of deep bidirectional transformers for language understanding. arXiv preprint arXiv:1810.04805  (2018)

\bibitem{drover2018can}
Drover, D., MV, R., Chen, C.H., Agrawal, A., Tyagi, A., Phuoc~Huynh, C.: Can 3d pose be learned from 2d projections alone? In: Proceedings of the European Conference on Computer Vision (ECCV) Workshops. pp.~0--0 (2018)

\bibitem{fang2018learning}
Fang, H.S., Xu, Y., Wang, W., Liu, X., Zhu, S.C.: Learning pose grammar to encode human body configuration for 3d pose estimation. In: Proceedings of the AAAI conference on artificial intelligence. vol.~32 (2018)

\bibitem{feichtenhofer2022masked}
Feichtenhofer, C., Li, Y., He, K., et~al.: Masked autoencoders as spatiotemporal learners. Advances in neural information processing systems  \textbf{35},  35946--35958 (2022)

\bibitem{gong2021poseaug}
Gong, K., Zhang, J., Feng, J.: Poseaug: A differentiable pose augmentation framework for 3d human pose estimation. In: Proceedings of the IEEE/CVF Conference on Computer Vision and Pattern Recognition. pp. 8575--8584 (2021)

\bibitem{hao2023divotrack}
Hao, S., Liu, P., Zhan, Y., Jin, K., Liu, Z., Song, M., Hwang, J.N., Wang, G.: Divotrack: A novel dataset and baseline method for cross-view multi-object tracking in diverse open scenes. arXiv preprint arXiv:2302.07676  (2023)

\bibitem{he2022masked}
He, K., Chen, X., Xie, S., Li, Y., Doll{\'a}r, P., Girshick, R.: Masked autoencoders are scalable vision learners. In: Proceedings of the IEEE/CVF Conference on Computer Vision and Pattern Recognition. pp. 16000--16009 (2022)

\bibitem{hu2021conditional}
Hu, W., Zhang, C., Zhan, F., Zhang, L., Wong, T.T.: Conditional directed graph convolution for 3d human pose estimation. In: Proceedings of the 29th ACM International Conference on Multimedia. pp. 602--611 (2021)

\bibitem{ionescu2013human3}
Ionescu, C., Papava, D., Olaru, V., Sminchisescu, C.: Human3. 6m: Large scale datasets and predictive methods for 3d human sensing in natural environments. IEEE transactions on pattern analysis and machine intelligence  \textbf{36}(7),  1325--1339 (2013)

\bibitem{li2019generating}
Li, C., Lee, G.H.: Generating multiple hypotheses for 3d human pose estimation with mixture density network. In: Proceedings of the IEEE/CVF conference on computer vision and pattern recognition. pp. 9887--9895 (2019)

\bibitem{li2021symbiotic}
Li, M., Chen, S., Chen, X., Zhang, Y., Wang, Y., Tian, Q.: Symbiotic graph neural networks for 3d skeleton-based human action recognition and motion prediction. IEEE Transactions on Pattern Analysis and Machine Intelligence  \textbf{44}(6),  3316--3333 (2021)

\bibitem{Li2020cascade3d}
Li, S., Ke, L., Pratama, K., Tai, Y.W., Tang, C.K., Cheng, K.T.: Cascaded deep monocular 3d human pose estimation with evolutionary training data. In: Proceedings of the IEEE/CVF conference on computer vision and pattern recognition. pp. 6173--6183 (2020)

\bibitem{li2022exploiting}
Li, W., Liu, H., Ding, R., Liu, M., Wang, P., Yang, W.: Exploiting temporal contexts with strided transformer for 3d human pose estimation. IEEE Transactions on Multimedia  (2022)

\bibitem{li2022mhformer}
Li, W., Liu, H., Tang, H., Wang, P., Van~Gool, L.: Mhformer: Multi-hypothesis transformer for 3d human pose estimation. In: Proceedings of the IEEE/CVF Conference on Computer Vision and Pattern Recognition. pp. 13147--13156 (2022)

\bibitem{lin2019trajectory}
Lin, J., Lee, G.H.: Trajectory space factorization for deep video-based 3d human pose estimation. arXiv preprint arXiv:1908.08289  (2019)

\bibitem{lin2014microsoft}
Lin, T.Y., Maire, M., Belongie, S., Hays, J., Perona, P., Ramanan, D., Doll{\'a}r, P., Zitnick, C.L.: Microsoft coco: Common objects in context. In: Computer Vision--ECCV 2014: 13th European Conference, Zurich, Switzerland, September 6-12, 2014, Proceedings, Part V 13. pp. 740--755. Springer (2014)

\bibitem{liu2020attention}
Liu, R., Shen, J., Wang, H., Chen, C., Cheung, S.c., Asari, V.: Attention mechanism exploits temporal contexts: Real-time 3d human pose reconstruction. In: Proceedings of the IEEE/CVF Conference on Computer Vision and Pattern Recognition. pp. 5064--5073 (2020)

\bibitem{loshchilov2017decoupled}
Loshchilov, I., Hutter, F.: Decoupled weight decay regularization. arXiv preprint arXiv:1711.05101  (2017)

\bibitem{lu2019vilbert}
Lu, J., Batra, D., Parikh, D., Lee, S.: Vilbert: Pretraining task-agnostic visiolinguistic representations for vision-and-language tasks. Advances in neural information processing systems  \textbf{32} (2019)

\bibitem{luo2020univl}
Luo, H., Ji, L., Shi, B., Huang, H., Duan, N., Li, T., Li, J., Bharti, T., Zhou, M.: Univl: A unified video and language pre-training model for multimodal understanding and generation. arXiv preprint arXiv:2002.06353  (2020)

\bibitem{mahmood2019amass}
Mahmood, N., Ghorbani, N., Troje, N.F., Pons-Moll, G., Black, M.J.: Amass: Archive of motion capture as surface shapes. In: Proceedings of the IEEE/CVF international conference on computer vision. pp. 5442--5451 (2019)

\bibitem{martinez2017simple}
Martinez, J., Hossain, R., Romero, J., Little, J.J.: A simple yet effective baseline for 3d human pose estimation. In: Proceedings of the IEEE international conference on computer vision. pp. 2640--2649 (2017)

\bibitem{mehta2017monocular}
Mehta, D., Rhodin, H., Casas, D., Fua, P., Sotnychenko, O., Xu, W., Theobalt, C.: Monocular 3d human pose estimation in the wild using improved cnn supervision. In: 2017 international conference on 3D vision (3DV). pp. 506--516. IEEE (2017)

\bibitem{paszke2019pytorch}
Paszke, A., Gross, S., Massa, F., Lerer, A., Bradbury, J., Chanan, G., Killeen, T., Lin, Z., Gimelshein, N., Antiga, L., et~al.: Pytorch: An imperative style, high-performance deep learning library. Advances in neural information processing systems  \textbf{32} (2019)

\bibitem{pavllo20193d}
Pavllo, D., Feichtenhofer, C., Grangier, D., Auli, M.: 3d human pose estimation in video with temporal convolutions and semi-supervised training. In: Proceedings of the IEEE/CVF conference on computer vision and pattern recognition. pp. 7753--7762 (2019)

\bibitem{radford2021learning}
Radford, A., Kim, J.W., Hallacy, C., Ramesh, A., Goh, G., Agarwal, S., Sastry, G., Askell, A., Mishkin, P., Clark, J., et~al.: Learning transferable visual models from natural language supervision. In: International conference on machine learning. pp. 8748--8763. PMLR (2021)

\bibitem{shan2022p}
Shan, W., Liu, Z., Zhang, X., Wang, S., Ma, S., Gao, W.: P-stmo: Pre-trained spatial temporal many-to-one model for 3d human pose estimation. In: Computer Vision--ECCV 2022: 17th European Conference, Tel Aviv, Israel, October 23--27, 2022, Proceedings, Part V. pp. 461--478. Springer (2022)

\bibitem{shan2021improving}
Shan, W., Lu, H., Wang, S., Zhang, X., Gao, W.: Improving robustness and accuracy via relative information encoding in 3d human pose estimation. In: Proceedings of the 29th ACM International Conference on Multimedia. pp. 3446--3454 (2021)

\bibitem{vaswani2017attention}
Vaswani, A., Shazeer, N., Parmar, N., Uszkoreit, J., Jones, L., Gomez, A.N., Kaiser, {\L}., Polosukhin, I.: Attention is all you need. Advances in neural information processing systems  \textbf{30} (2017)

\bibitem{xu2020deep}
Xu, J., Yu, Z., Ni, B., Yang, J., Yang, X., Zhang, W.: Deep kinematics analysis for monocular 3d human pose estimation. In: Proceedings of the IEEE/CVF Conference on computer vision and Pattern recognition. pp. 899--908 (2020)

\bibitem{xu2023multimodal}
Xu, P., Zhu, X., Clifton, D.A.: Multimodal learning with transformers: A survey. IEEE Transactions on Pattern Analysis and Machine Intelligence  (2023)

\bibitem{yan2018spatial}
Yan, S., Xiong, Y., Lin, D.: Spatial temporal graph convolutional networks for skeleton-based action recognition. In: Proceedings of the AAAI conference on artificial intelligence. vol.~32 (2018)

\bibitem{yang2021feedback}
Yang, H., Yan, D., Zhang, L., Sun, Y., Li, D., Maybank, S.J.: Feedback graph convolutional network for skeleton-based action recognition. IEEE Transactions on Image Processing  \textbf{31},  164--175 (2021)

\bibitem{zeng2020srnet}
Zeng, A., Sun, X., Huang, F., Liu, M., Xu, Q., Lin, S.: Srnet: Improving generalization in 3d human pose estimation with a split-and-recombine approach. In: Computer Vision--ECCV 2020: 16th European Conference, Glasgow, UK, August 23--28, 2020, Proceedings, Part XIV 16. pp. 507--523. Springer (2020)

\bibitem{Zhang2022MIXSTE}
Zhang, J., Tu, Z., Yang, J., Chen, Y., Yuan, J.: Mixste: Seq2seq mixed spatio-temporal encoder for 3d human pose estimation in video. In: Proceedings of the IEEE/CVF Conference on Computer Vision and Pattern Recognition. pp. 13232--13242 (2022)

\bibitem{zhao2023poseformerv2}
Zhao, Q., Zheng, C., Liu, M., Wang, P., Chen, C.: Poseformerv2: Exploring frequency domain for efficient and robust 3d human pose estimation. In: Proceedings of the IEEE/CVF Conference on Computer Vision and Pattern Recognition. pp. 8877--8886 (2023)

\bibitem{zheng20213d}
Zheng, C., Zhu, S., Mendieta, M., Yang, T., Chen, C., Ding, Z.: 3d human pose estimation with spatial and temporal transformers. In: Proceedings of the IEEE/CVF International Conference on Computer Vision. pp. 11656--11665 (2021)

\end{thebibliography}
\section*{Appendix}

\subsection*{A. What is the Best Structure Design? }
We compare the performance of different layers of transformer in shared layers because shared layers occupy the majority of the m model's parameters. The result is shown in Table~\ref{tab:abl_model}. It demonstrates that more frames as input achieves higher performance, and 3D Decoder in \stage{2} also improves our model's 3D HPE ability. With the increase of shared layers, the performance first improves, then drops. Shared layers in \stage{1} learn 2D and 3D features in the same feature space, so compared to 2, and 3 layers, 4 shared layers can learn better common features during pre-training. However, 5 shared layers seem to cause over-fitting.
\renewcommand{\thetable}{A}
\begin{table}[!h]
    \centering
    \caption{Performance for different model designs using CPN 2D as input on Human3.6M benchmark in terms of MPJPE and P-MPJPE metric.}
    \resizebox{0.6\linewidth}{!}{
    \begin{tabular}{c|>{\centering\arraybackslash}p{1.2cm}>{\centering\arraybackslash}p{1.8cm}|cc}
    \toprule
    Shared Layers & Frames & 3D Decoder & MPJPE~($\downarrow$) & P-MPJPE~($\downarrow$)       \\ 
    \midrule
    4      & 243    &  \stage{1}   & 45.4     & 35.9     \\
    \midrule
    4      & 81     & \stage{2}  & 43.6     & 35.0       \\
    \midrule
    2      & 243    &  \stage{2}     &  43.2        &   35.1                       \\
    3      & 243    &  \stage{2}     &  42.8        &     34.5                     \\
    4      & 243     &  \stage{2}    & \textbf{42.3}     & \textbf{34.4}  \\
    5      & 243     &  \stage{2}    &  42.7        &      34.4          \\
    \bottomrule
    \end{tabular}
    }
    \label{tab:abl_model}
\end{table}

\subsection*{B. Does Mask Sampling Strategies Matter?}
\label{mask}
We conduct ablation studies on three different mask sampling strategies as shown in Figure~\ref{fig:mask} including: 1) spatial masking or called tube masking, in which certain joints are masked over all the frames; 2) temporal masking, in which all the joints in certain frames are masked; and 3) spatiotemporal masking, which is our final choice, in which masked joints are picked frame by frame randomly. We conduct comparison experiments under the same masking ratio of $71.8\%$ in total as shown in Table~\ref{tab:mask}. Spatiotemporal masking outperforms the other two mask sampling strategies, leading to better relation modeling both spatially and temporally.

\renewcommand{\thefigure}{B}
\begin{figure}[!t]
    \centering
    \includegraphics[width=0.55\linewidth]{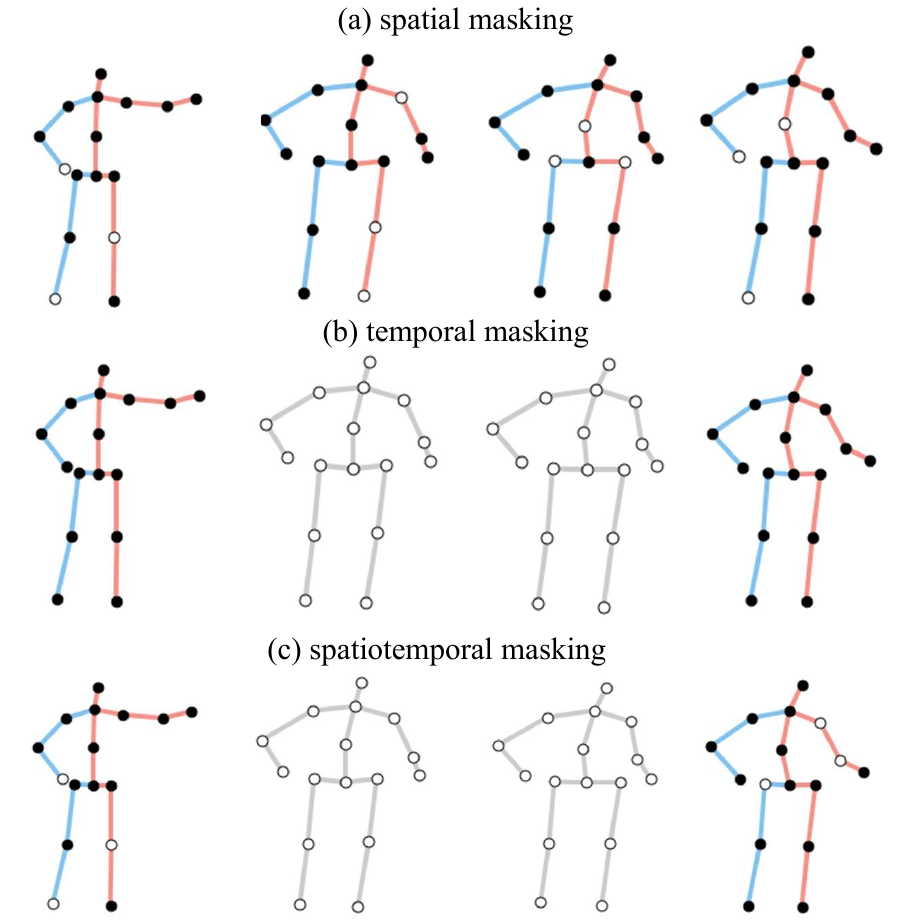}
    \caption{Illustration of three different mask sampling strategies. The joints in white/gray color represent a masked one and the joints in black color represent the unmasked one.}
    \label{fig:mask}
\end{figure}
\renewcommand{\thetable}{B}
\begin{table}[t]
\centering
\caption{Performance under different three mask sampling strategies on Human3.6M and MPI-INF-3DHP in 27 frames setting.}
\resizebox{0.7\linewidth}{!}
{
\begin{tabular}{l|cc|ccc}
\toprule
\multirow{2}{*}{Mask} & \multicolumn{2}{c|}{Human3.6M} & \multicolumn{3}{c}{MPI-INF-3DHP} \\
& MPJPE~($\downarrow$) & PA-MPJPE~($\downarrow$) & PCK~($\uparrow$) & AUC~($\uparrow$) & MPJPE~($\downarrow$)\\
\midrule
Spatial & 47.3 & 37.5 & 97.4 & 71.7 & 42.4\\
Temporal & 47.7 & 38.0 & 97.3 & 71.1 &  43.1\\
Spatio-temporal & \textbf{46.7} & \textbf{97.9} & \textbf{97.7} & \textbf{72.2} & \textbf{41.4}\\
\bottomrule
\end{tabular}
}
\label{tab:mask}
\end{table}

\subsection*{C. What is the Best Masking Ratio in \mpm?}
The detailed result of the grid search of optimal mask ratio in Figure~6 is shown in Table~\ref{tab:maskres}.

\renewcommand{\thetable}{C}
\begin{table}[!h]
    \centering
     \caption{Grid search of masking number of joints $r_s$ and ratio $r_t$ for spatiotemporal masking. We report the performance over MPJPE  on Human3.6M. The optimal ratios are $r_s^* = 5/17 $ in spatial and $r_t^* = 60\%$ in temporal leads to $46.7$ mm of MPJPE.}
    \resizebox{0.4\linewidth}{!}{
    \begin{tabular}
    {c|cccccc}
    \toprule
      $r_s$ v.s. $r_t$   & 0.4 & 0.5 & 0.6 & 0.7 & 0.8\\
    \midrule
       2 & 47.2& 46.9 &47.0 &47.1   & 47.2 \\
       3 & 47.5& 47.4 &47.2 &47.6   & 47.3\\ 
       5 & \underline{46.9}& 47.3 &\textbf{46.7} &47.7   & 47.2\\ 
       7 & 47.6& 47.2 &47.5 &47.2   & 47.2 \\ 
       9 & 47.3& 47.3 &47.5 &47.6   & 47.3 \\
    \bottomrule
    \end{tabular}
    }
    \label{tab:maskres}
\end{table}

\subsection*{D. Comparison with other methods in numbers of parameters}
As shown in Table~\ref{tab:parmcompare}, our method performs well in terms of parameter quantity and performance.
\vspace{-10pt}
\renewcommand{\thetable}{D}
\begin{table}[!h]
\scriptsize
    \centering
    \caption{Comparation with other methods in numbers of parameters.}
    \resizebox{0.65\linewidth}{!}{
    \begin{tabular}
    {c|c|ccc}
    \toprule
      Method & \#Frames  & Param. [M] & FLOPs [M] & MPJPE [mm] \\
    \midrule
       Chen \etal \cite{chen2021anatomy} & 81 & 58.1& \underline{116} & 32.3 \\ 
       Zhang \etal \cite{Zhang2022MIXSTE}& 243 & 33.8 & 156735 & \textbf{21.6} \\
       Zhang \etal \cite{Zhang2022MIXSTE}& 81 & 33.7 & 52245 & \underline{25.9} \\
        Li \etal \cite{li2022mhformer} &  351  &18.9 & 1030 & 30.5 \\
          Pavllo \etal \cite{pavllo20193d}  & 243 & 17.0&  \textbf{34} & 37.8\\ 
         Zheng \etal \cite{zheng20213d} & 81 &9.6& 1358 & 31.3 \\
       Shan \etal \cite{shan2022p}&  243 &  \textbf{6.7} & 1737 & 29.3 \\  
       MPM(ours) & 243 &\underline{7.8} & 2387 & 27.2 \\ 
       
    \bottomrule
    \end{tabular}
    }
    \label{tab:parmcompare}
\end{table}

\renewcommand{\thetable}{E}
\renewcommand{\thefigure}{E}
\begin{figure}[!h] 
\vspace{-5pt}
  \begin{minipage}[c]{0.4\textwidth} 
    \centering
    \resizebox{\linewidth}{!}
        {
        \begin{tabular}{c|ccc|c}
            \toprule
            Setting & Human3.6M & MPI-INF-3DHP & AMASS & \# Samples \\
            \midrule
            Small & $\checkmark_{\text{half}}$ & $\checkmark_{\text{quarter}}$ & & 4.17 M\\
            Base & $\checkmark$ & $\checkmark_{\text{half}}$ & & 8.34 M\\
            Large & $\checkmark$ & $\checkmark$ & $\checkmark$ & 18.53 M\\
            \bottomrule
            \end{tabular}
        }
    
  \makeatletter\def\@captype{table}\makeatother
    \caption{Three different training data scales. $\checkmark_{\text{half}}$ and $\checkmark_{\text{quarter}}$ denotes using half and quarter of the data.} 
    \label{tab:pretrain}
  \end{minipage} 
    \hspace{20pt} %
   \begin{minipage}[c]{0.5\textwidth} 
    \centering 
    \includegraphics[width=1.1\textwidth]{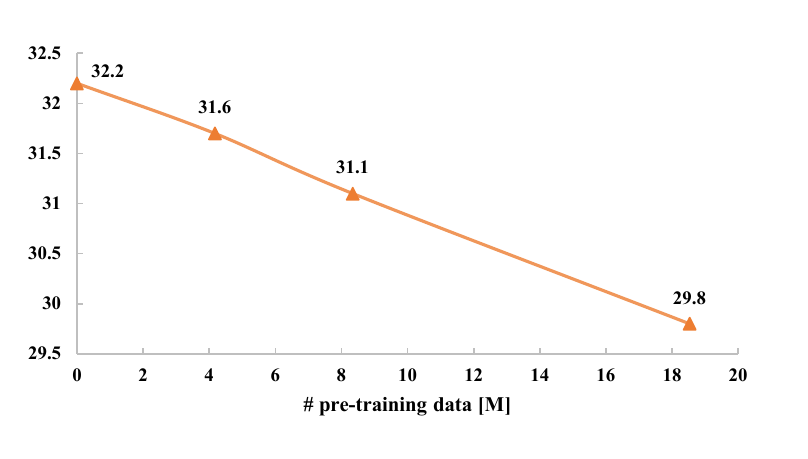} 
     \caption{Performance under different pre-training data scales shown in Table~\ref{tab:pretrain} under Protocol \#1 on MPI-INF-3DHP.}
    \label{fig:pretrain}
  \end{minipage}%
  \vspace{-10pt}
\end{figure}

\vspace{-20pt}
\subsection*{E. Data Scaling Law of \mpm.}
We claim that a good pre-training framework could benefit from larger pre-training data. To verify this kind of data scaling law in \mpm, we pre-train our network under three different training data scales as shown in Table~\ref{tab:pretrain}. Figure~\ref{fig:pretrain} demonstrates the plot of performance under different pre-training data scales. We show that our \mpm~could benefit from large training data in \stage{1}.

\end{document}